\documentclass[runningheads]{llncs}
\usepackage{graphicx}
\usepackage{bbm}
\usepackage{tikz}
\usepackage{float}
\usepackage{comment}
\usepackage{amsmath,amssymb} 
\usepackage{color}
\usepackage{graphicx}
\usepackage{sidecap}
\usepackage{multirow} 
\usepackage{wrapfig}
\usepackage[misc,geometry]{ifsym}
\newcommand{\etc}{\textit{etc}}
\newcommand{\eg}{\textit{e}.\textit{g}.}
\newcommand{\ie}{\textit{i}.\textit{e}.}
\newcommand{\etal}{\textit{et al}. }

\usepackage[accsupp]{axessibility}  

\graphicspath{{images/}}
\begin{document}
\pagestyle{headings}
\mainmatter
\def\ECCVSubNumber{2766}  

\title{Generalizable Person Re-Identification via Viewpoint Alignment and Fusion} 
\author{Bingliang Jiao\inst{1,2,3,5} \and
	Lingqiao Liu\inst{4} \and
	Liying Gao\inst{1,2,3} \and Guosheng Lin\inst{5} \and Ruiqi Wu\inst{1,2,3} \and Shizhou Zhang\inst{1,3} \and Peng Wang\inst{1,2,3}  \and Yanning Zhang\inst{1,3}}
\institute{School of Computer Science, Northwestern Polytechnical University, China \and
	Ningbo Institute, Northwestern Polytechnical University, China \and
	National Engineering Laboratory for Integrated Aero-Space-Ground-Ocean, China
	\and The University of Adelaide, Australia \and Nanyang Technological University, Singapore
}
\maketitle


\begin{abstract}
In the current person Re-identification (ReID) methods, most domain generalization works focus on dealing with style differences between domains while largely ignoring unpredictable camera view change, which we identify as another major factor leading to a poor generalization of ReID methods. To tackle the viewpoint change, this work proposes to use a 3D dense pose estimation model and a texture mapping module to map the pedestrian images to canonical view images. Due to the imperfection of the texture mapping module, the canonical view images may lose the discriminative detail clues from the original images, and thus directly using them for ReID will inevitably result in poor performance. To handle this issue, we propose to fuse the original image and canonical view image via a transformer-based module. The key insight of this design is that the cross-attention mechanism in the transformer could be an ideal solution to align the discriminative texture clues from the original image with the canonical view image, which could compensate for the low-quality texture information of the canonical view image. Through extensive experiments, we show that our method can lead to superior performance over the existing approaches in various evaluation settings. 
\end{abstract}

\section{Introduction}
Person re-identification (ReID) is a task to match identical persons across different camera views. 
Due to its applications in intelligent surveillance and tracking, ReID has drawn a growing amount of interest.
Recently, supervised ReID has achieved encouraging performance~\cite{ye2021deep,zhang2021hat,zhang2020relation,sun2018beyond}.
However, these methods always suffer significant degradation when deployed into a new scenario (domain), which limits the real-world application of ReID algorithms.


\begin{figure*}[t]
	\begin{center}
		\includegraphics[width=0.6\linewidth]{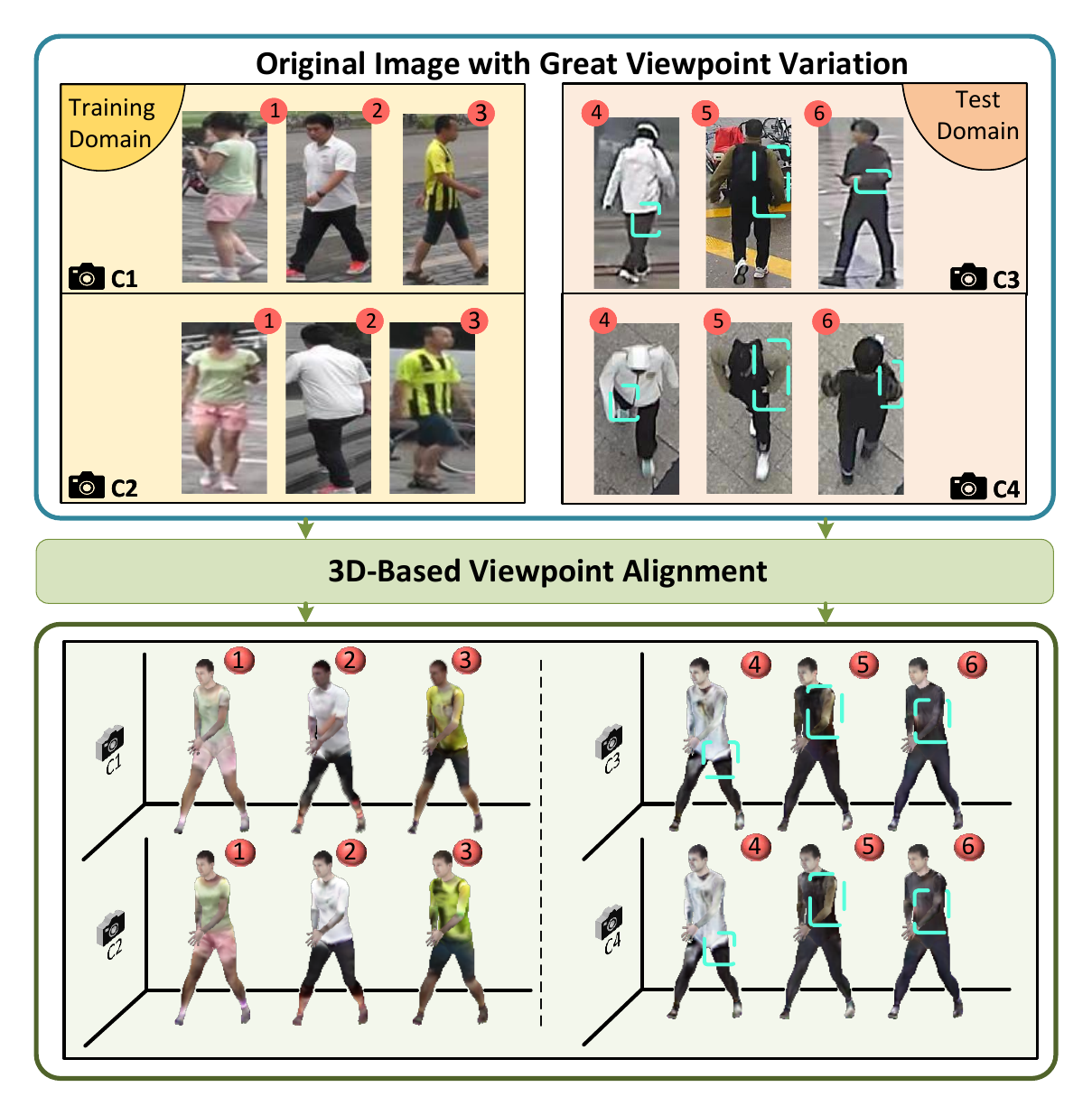}
	\end{center}
    \vspace{-4mm}
	\caption{The sketch of 3D-based viewpoint alignment. As shown in the upper part, the unseen camera views could cause drastic pattern and geometry deformation to pedestrian objects (marked by blue bounding boxes), which introduces a great challenge to the generalization of ReID model. Hence, in this work, we propose to address this challenge by using 3D space as a bridge to calibrate the viewpoint of pedestrians from arbitrary domains. 
		}
	\label{fig:Figure1}
	\vspace{-4mm}
\end{figure*}

In recent years, many relevant methods have been proposed to address this issue.
Generally, these methods could be grouped into two types, namely,  domain adaptation ReID and domain generalization ReID (DG-ReID).
For the former~\cite{liu2019adaptive,tang2020cgan,zhai2020ad,zheng2021online}, they assume that partial target domain unlabelled data are accessible during the training stage. However, this is not always available in real applications.
DG-ReID does not make this assumption but aims to learn a ReID model that can generalize to arbitrary domains without fine-tuning, and thus it is less restrictive but also more challenging. 

Among existing DG-ReID works, the style divergence between domains is widely considered a major factor limiting the generalization of ReID models, and various methods~\cite{jia2019frustratingly,jin2020style,choi2021meta,jiao2022dynamically} have been proposed to address the degradation caused by style divergence. However, another major factor affecting domain generalization: camera-view variation, has rarely been addressed in the literature. 
Generally, the camera-view variation could cause drastic pattern and geometry deformation to pedestrian objects. For the domain generalization task, this issue could become much more challenging since the deformations introduced by novel camera views are unpredictable. 
As shown in Figure~\ref{fig:Figure1}, the novel camera view in the test domains could cause significant and unseen appearance deformation to the pedestrian patterns (marked by blue bounding boxes).


This work aims to fill the gap by explicitly addressing the camera-view variation. We propose a 3D-based Viewpoint Alignment Network (3D-VAN) to alleviate the impact of viewpoint divergence across domains.
The basic idea of our 3D-VAN is illustrated in Figure~\ref{fig:Figure1}.
Specifically, our 3D-VAN uses 2D-to-3D reconstruction methods to map 2D pedestrian images into the 3D space, from which we could generate canonical view images by utilizing a unified posture and viewpoint to obtain projections from the reconstructed 3D models.
In practice, to reconstruct 3D human models, we employ a 3D dense pose estimation model RSC-Net~\cite{xu20213d} and a texture mapping module Texformer~\cite{xu20213dtext} to predict the latent parameter and texture for the 3D deformable human model SMPL~\cite{loper2015smpl}.
However, the existing 3D texture predictor does not perform well in the cross-domain setting. We thus propose to incorporate an additional normalization module to make it more robust.

Meanwhile, due to the poor quality of surveillance images and the imperfection of single-view 3D reconstruction, the canonical view images inevitably lose some discriminative clues from the original images.
We thus cannot directly use the generated canonical view images for identification, as it could lead to unsatisfactory performance. 
To address this issue, in this work, we propose to fuse the original image to the canonical view image to compensate for the lost details. Such a fusion can be challenging since the original image and canonical view image have different views. Ideally, the fusion model needs to first align the key parts of the person, \eg, head, shoulder, or legs, from both images and then combine the visual clues for the same part. 
In our 3D-VAN, we propose a transformer-based feature fusion module to accomplish this goal.
Our key insight is that the attention mechanism inside a transformer could be suited for modeling the ``aligning-and-fusion'' operation as described above.
In our design, we also further modify the structure of the transformer to make it more suited for our fusion task.
With the fusion module, our 3D-VAN could implicitly compensate for the low-quality texture information of the canonical view images from the original images and achieve better feature presentations for ReID.

Through extensive experiments, the superiority of our designed modules has been validated. Besides, under three mainstream DG-ReID settings, our 3D-VAN outperforms state-of-the-art approaches.
To summarize, our main contributions are three folds:
\begin{itemize}
\item In this work, we propose a 3D-VAN model for DG-ReID, which is a pioneering work on enhancing the generalization capability of the ReID system by explicitly addressing the viewpoint variation issue in cross-domain settings.
\item  In our 3D-VAN, we use the 3D space as a bridge to calibrate the viewpoint of pedestrians from arbitrary domains. We validate the feasibility and improve the robustness of existing 2D-to-3D reconstruction models for this purpose.
\item Meanwhile, we also design a transformer-based feature fusion module, which aligns and fuses the texture clues from original images to canonical view images to compensate for the lost details caused by imperfect 3D reconstruction methods.
\end{itemize}

\section{Related Work}

\noindent{\bf {Domain Generalization Person Re-Identification.}} 
To alleviate the degradation of existing ReID methods when facing novel domains, song~\etal~\cite{song2019generalizable} propose the domain generalization person re-identification (DG-ReID) task.
The DG-ReID aims to train a ReID system that can work well on arbitrary domains.
As suggested in~\cite{pan2018two,jin2020style,choi2021meta,jiao2022dynamically}, the style divergence among different domains is a major factor that affects the generalization of ReID models.
Therefore, many  works~\cite{jia2019frustratingly,jin2020style,choi2021meta, jiao2022dynamically} propose to use instance normalization technology to alleviate the degradation caused by style divergence.
These methods over-focus on the style alignment but ignore the camera view variation, which we identify as another salient factor limiting the generalization of ReID methods.
Therefore, in this work, we propose the 3D-VAN, which explicitly addresses the viewpoint change in the cross-domain setting to facilitate the generalization of ReID models.

\noindent{\bf{3D Human Reconstruction.}} Recently, 3D object reconstruction has attracted more and more interest due to its commercial value. 
The 3D object reconstruction aims to learn a transformation projection to map target objects from 2D images into 3D space.
Here, we revisit three mainstream 3D reconstruction methods and discuss the feasibility of employing them in the ReID task.
Firstly, Mildenhall~\etal~\cite{mildenhall2020nerf} propose a NeRF model which can learn the 3D information of a target object from its images captured under multiple discrete camera views.
Since we are only allowed to access a single view of the target pedestrian instance in the ReID task, the NeRF and its variants~\cite{gao2022mps,chen2021mvsnerf,martin2021nerf} thus cannot be employed. 
Secondly, Saito~\etal~\cite{saito2019pifu} propose a PIFu model which can reconstruct the 3D human surface from a single image.
The PIFu is devoted to directly predicting all 3D vertice coordinates from a 2D image without any prior, which highly relies on the image quality. 
Through our evaluation, we find that the PIFu can hardly fit low-resolution inputs and thus cannot be used to tackle surveillance imagery in the ReID task.
Thirdly, there are also some effective parametric 3D human models~\cite{anguelov2005scape,loper2015smpl}. 
For instance, with a set of pose parameters and shape parameters describing the target pedestrian, the SMPL~\cite{loper2015smpl} can reconstruct the corresponding 3D human mesh leverage to its prior knowledge.
Based on SMPL, some effective 3D prediction methods~\cite{xu20213d, xu20213dtext} have also been proposed, which can predict necessary potential parameters and textures for SMPL from single-frame surveillance imagery.
Therefore, in this work, we choose SMPL~\cite{loper2015smpl} and relevant 3D prediction methods~\cite{xu20213d, xu20213dtext} to reconstruct 3D pedestrian models.


\section{Method}

\begin{figure*}[t]
	\begin{center}
		\includegraphics[width=0.95\linewidth]{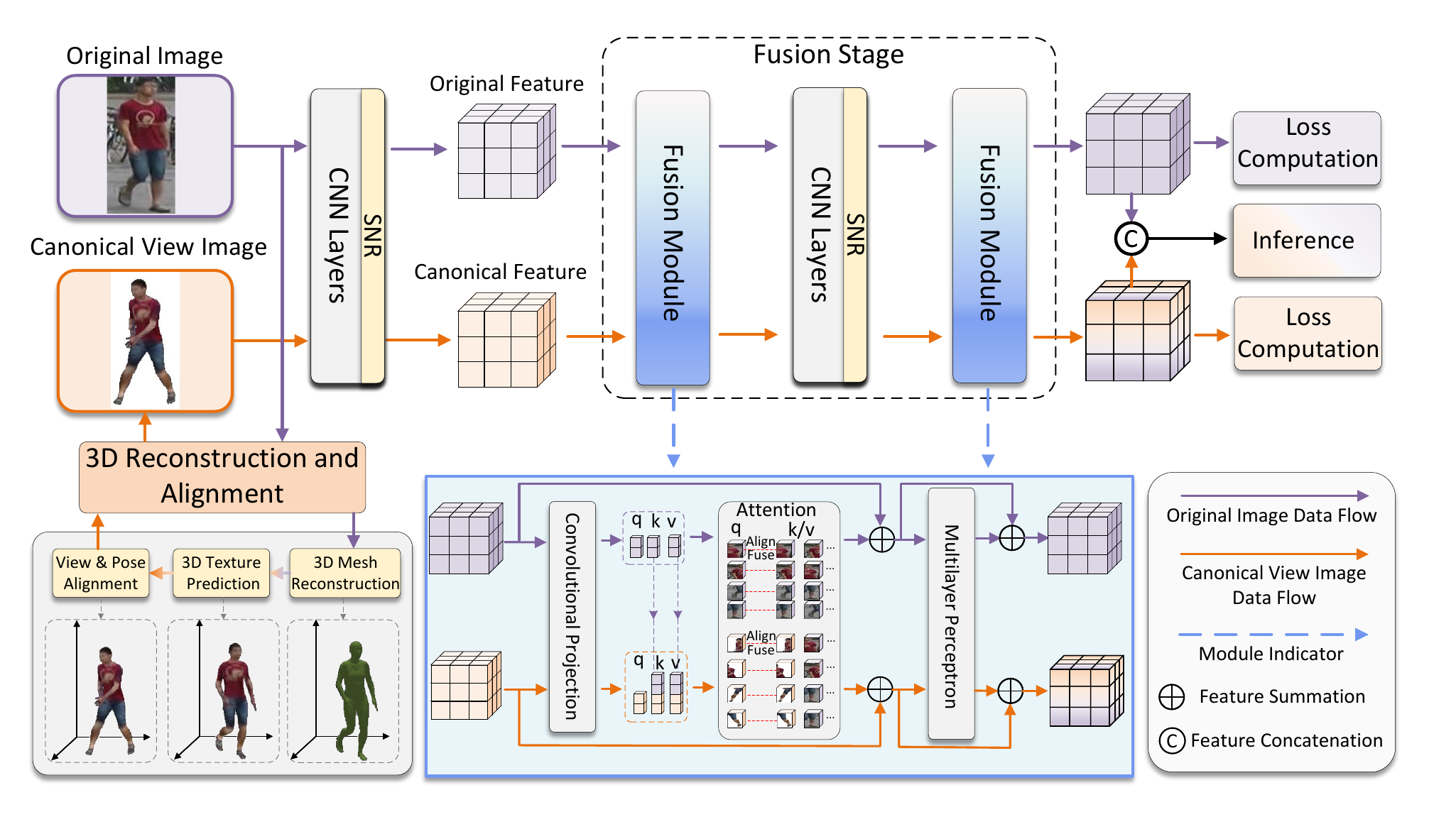}
	\end{center}
    \vspace{-8mm}
	\caption{
		The overall framework of our 3D-VAN.
		Our 3D-VAN extracts original image features and canonical view image features in parallel and realizes the alignment and fusion between them through a carefully designed transformer-based feature fusion module. The SNR~\cite{jin2020style} is an adaptive instance normalization module, which is employed to filter out style interference.
		}
	\label{fig:structure}
	\vspace{-4mm}
\end{figure*}

In this section, we introduce our proposed 3D-based viewpoint alignment network (3D-VAN).
The overall framework is shown in Figure~\ref{fig:structure}. There are two main components in our model: (1) method to map 2D images to 3D and obtain the canonical view images. (2) a feature fusion module that combines clues from both the canonical view image and the original image. Those two components are introduced in the following two sections. More illustrations are provided in the Appendix, \eg,  Figure A in the Appendix intuitively visualizes the process of the first component.

\subsection{3D Human Reconstruction and Alignment}\label{sec:rec} 

\noindent{\bf{3D Pedestrian Mesh Reconstruction.}} In this work, we represent the 3D human body with the 3D morphable human model SMPL~\cite{loper2015smpl}. The SMPL is a parametric model which constructs 3D pedestrian mesh with a group of shape and pose parameters denoted as $\beta$ and $\theta$, respectively.
To be specific, SMPL describes the human body shape using a set of basis vectors that are learned from a large number of 3D body scans through principal component analysis. The shape parameter $\beta \in \mathbb{R}^{10}$ represents the coefficients of shape basis vectors. 
Meanwhile, the SMPL uses a skeleton structure to describe the body pose with $N=24$ joints.
The $\theta \in \mathbb{R}^{3 \times N}$ denotes relative
rotation between adjacent joints and the global orientation of 3D models.
Formally, the SMPL can be written as:
\begin{equation}
    M = W(\beta, \theta),
\end{equation}
where $M \in \mathbb{R}^{3 \times K}$ is the predicted 3D mesh with $K$ vertices; the $W(\cdot)$ is the function of the SMPL model.

In this work, we utilize the pre-trained 3D dense pose estimation model  RSC-Net~\cite{xu20213d} to predict the shape parameter $\beta$ and pose parameter $\theta$ from every input image.



\noindent{\bf{Texture Prediction for 3D Pedestrian Mesh.}}
In this step, we employ a texture mapping module Texformer~\cite{xu20213dtext} to map the texture details of the original image to the constructed 3D mesh.
Unfortunately, the original Texformer is susceptible to domain variation and thus may not perform well in new scenarios.
To tackle this limitation, we make two ingenious refinements to Texformer to enhance its robustness to domain changes.
In the rest of this manuscript, we refer to the refined version as R-Texformer.
For easy understanding, in Figure B in the Appendix, we give a sketch of the R-Texformer.

Specifically, we modify Texformer in both network structure and training strategy.
For structure, we insert several adaptive instance normalization layers SNR~\cite{jin2020style} into the feature extraction module of Texformer to alleviate the degradation caused by the feature distribution divergence across domains.
Regarding training strategy, we propose a content-biased training strategy to ensure that Texformer makes predictions based on pedestrian-relevant content rather than style clues. 
In the following, we elaborate on the content-biased training strategy.

During the training stage, we first employ a pre-trained style-transferring model AdaIN~\cite{huang2017arbitrary} to randomize the style of each input image $I$.
AdaIN uses the auto-encoder framework, and we following~\cite{luo2020adversarial} to inject random noise into its extracted hidden features to generate style-randomized images.
After that, we train the Texformer with a content-biased loss to enforce it to make predictions based on the content information.
Formally, within a mini-batch, we sample three images, \ie, an original image $I_{p}$, its style-randomized result $I_{a}$, and another style-randomized image $I_{n}$ that shares the same random style with $I_{a}$ but has a different identity. The content-biased loss could be written as,
\begin{equation}
    L_c=max(0,\;\tau +d(\varphi(I_{a}),\varphi(I_{p}))-d(\varphi(I_{a}),\varphi(I_{n})))
\end{equation}
where the $\tau$ is a margin; the $\varphi$ denotes the Texformer model; the $d(\cdot)$ is the distance between predicted textures, which is the euclidean distance between the features of predicted texture maps extracted by a pre-trained ResNet-50~\cite{he2016deep}.
In this way, we can train the Texformer to focus on the consistent content information between $(I_{a}$,$I_{p})$ while ignoring the consistent style information between $(I_{a}$,$I_{n})$.
For more technical details of these refinements, please refer to Section B in the Appendix.

\noindent{\bf{3D Posture and Viewpoint Alignment.}}
Finally, we align the viewpoint of all 3D meshes and map them into canonical view images.
In this step, we also align the posture of 3D meshes, which can make key human parts, \eg, head, shoulder, or legs, projected in fixed areas of the canonical view images.
This explicit human part alignment operation could help the ReID model to capture subtle discrepancies between key parts of different pedestrians.
Here, we accomplish this goal by unifying the pose parameter $\theta$ of all SMPL meshes.
After that, we render the aligned 3D model back to 2D space from several canonical observing views.
In this work, we select four canonical camera views, namely, forward, backward, left, and right views, to avoid the loss of textural detail information.  Formally, given a camera parameter $\delta \in \mathbb{R}^{3}$ which defines the 3D translation of the observing view, the process of 3D to 2D rendering could be written as, 
\vspace{-1mm}
\begin{equation}
    J = f_{p}(M, \delta ),
\end{equation}
\vspace{0.5mm}where the $f_{p}(\cdot)$ is the projection function in~\cite{hartley2003multiple}; the $M$ is the set of 3D pedestrian vertex locations; the $J$ is the locations of rendered 2D keypoints.

\subsection{Transformer-Based Feature Fusion Module}
Due to the imperfection of 2D-to-3D reconstruction, the canonical view images generated from single-view surveillance images inevitably lose discriminative texture clues, and thus directly using them for ReID could result in poor performance.
Hence, in this work, we propose to fuse the original image and canonical view image in the feature space to compensate for the lost details. The main challenge of this fusion module is that the canonical view image and original image might not be in the same view or pose. Thus it is crucial to have an alignment mechanism inside the fusion module such that the visual clues from the same part, \eg, head, leg, \etc, can be combined together. In this paper, we propose a transformer-based module to achieve this goal. A transformer-based module could be an ideal solution since the attention mechanism inside the transformer can simultaneously model both the alignment (through affinity) and fusion (through 
aggregation) operations. 

Specifically, in our 3D-VAN, we first use a shared ResNet-50~\cite{he2016deep} model to extract the features of the original image and canonical view image belonging to the same pedestrian instance.
In this step, we find that there exists a distribution gap between features of real original images and features of synthetic images, which could limit the alignment between them.
To address this issue, in our 3D-PAN, we apply separate Batch Normalization (BN)~\cite{ioffe2015batch} to features from different types of input images.
Taking the features of canonical view images $F_{c}$ in the current mini-batch as an example, the BN employed in our 3D-VAN could be written as,
\begin{align}
&\text{Standardization:} ~~\widehat F_{c} = \frac{F_{c}-\mu(F_{c})}{\sigma(F_{c})}, \nonumber \\
&\text{Transformation:} ~~\widetilde F_{c} = \gamma_{c} \widehat F_{c} + \beta_{c},
\label{Eq:Norm}
\end{align} 
where the $\widetilde F_{c}$ indicates the normalized canonical view image features; the $\mu(\cdot)$ and $\sigma(\cdot)$ are functions to calculate channel-wise mean and standard deviation; the $\gamma_{c}$ and $\beta_{c}$ are affine parameters specifically learned for canonical view features.
By respectively calibrating features of original images and canonical view images, the distribution conflict between them could be alleviated.
In this way, we could help the network to embed them into a unified feature space and thus facilitate feature alignment.

After extracting the features of canonical view images and original images, we then send them into our feature fusion module.
Formally, our feature fusion module adopts the Transformer structure~\cite{vaswani2017attention}, the sketch of which is given in Figure~\ref{fig:structure}.
Practically, to preserve the spatial information of input features, in our feature fusion module, we follow CVT~\cite{wu2021cvt} to project input features into query, key, and value with convolution operations.
The process of feature projection can be formulated as follows:
\begin{align} 
Q/K/V = \mathrm{Flatten}(\mathrm{Conv2d}(F)),
\label{Eq:QKV}
\end{align}

\noindent where the $F$ denotes the input features; the $Q/K/V$ represents the query/key/value vector; $\mathrm{Conv2d}(\cdot)$ represents a convolutional layer.
Then we use a cross-attention operation to align and fuse the features of cross-view inputs.
In this operation, we propose an asymmetric access restriction, in which we only allow the canonical view image features to access the original image features while forbidding the opposite information flow.
The purpose of the asymmetric design is to avoid distorted textural information transferred from synthetic image features to original image features.
Formally, the cross-attention we apply can be written as,
\begin{align} 
&\text{Affinity:}~~ A_{c} = \delta (Q_{c} [K_{c}, K_{o}]^{T}),~A_{o} = \delta (Q_{o}K_{o}^{T}) \nonumber\\
&\text{Aggregation:}~~ F_{c}^{a} = A_c[V_c,V_o],~~~F_{o}^{a} = A_oV_o
\label{Eq:attn}
\end{align}
where the subscript $c/o$ indicates the current element belonging to canonical view image ($c$) or original image ($o$); the $F^{a}$ is the output features; 
the $\delta$ indicates the Softmax function; 
the $[\cdot]$ is feature concatenation in the flattened spatial dimension.
Practically, we follow multi-head attention~\cite{vaswani2017attention} to split the $Q/K/V$ in channel dimension into multi-head groups and independently apply attention on each group according to Equation~\ref{Eq:attn}.
A multi-layer perceptron is also inserted after the attention block. 
By aligning and fusing the textural information from original image features to canonical view image features, the lost details of canonical view image features could be implicitly compensated.
In Figure~\ref{fig:visualization}, we give some intuitive cases to show how our fusion module aligns and fuses the information of textural details between original images and canonical view images.

\noindent{\bf The Deployment of Feature Fusion Module.} In our 3D-VAN network, we use the ResNet-50 as the backbone model. The feature fusion module is inserted after the 3rd and 4th layers of ResNet-50, respectively.
Meanwhile, to alleviate the interference caused by style variation, in our 3D-VAN, the SNR~\cite{jin2020style} module is inserted after each layer of ResNet-50 (before our feature fusion module).

\noindent{\bf Training and Test Procedure.}
During the training stage, the original image features and canonical view image features are respectively used for ID classification loss and triplet loss calculation.
In addition, in the last fusion module, we employ a learnable vector to serve as the class token~\cite{wu2021cvt}, which could adaptively aggregate features from both original image tokens and canonical view image tokens. 
The class token is used to calculate ID classification loss, as well.
The purpose of using the class token is to make the canonical view image features and original image features to be more compatible. 
The overall objective function of our 3D-VAN network can thus be formulated as:

\begin{align} 
L=L_{o}^{Tri.} + L_{c}^{Tri.} +  L_{o}^{ID} + L_{c}^{ID} + L_{cls}^{ID}, 
\label{Eq:Loss}
\end{align}

\begin{table*}
	\begin{center}
\caption{The comparison between our 3D-VAN and other state-of-the-art domain generalization person ReID methods under multiple-source domain protocols, namely, Protocol-1 (P-1) and Protocol-2 (P-1). It can be found that our 3D-VAN outperforms all compared methods under both protocols.}
	\label{tab:multiple}
	\vspace{-1mm}
	\resizebox{0.85\textwidth}{!}{ 
			\begin{tabular}{c| c | c | c c c c c c| c c}
				\hline
		         \multirow{2}*{~} & \multirow{2}*{Method} & \multirow{2}*{Reference} &  \multicolumn{2}{c}{GRID} &  \multicolumn{2}{c}{VIPeR} & \multicolumn{2}{c|}{i-LIDS} & \multicolumn{2}{c}{Average} \\
				 ~ & & & mAP & CMC-1 & mAP & CMC-1 & mAP & CMC-1  & mAP & CMC-1 \\
				\hline \hline
				& DIMN~\cite{song2019generalizable} & CVPR2019 & $41.1$ & $29.3$ & $60.1$ & $51.2$ & $78.4$ & $70.2$ & $59.9$ & $50.2$ \\ 	
				& DMG-Net~\cite{bai2021person30k} & CVPR2021 & $56.6$ & $51.0$ & $60.4$ & $53.9$ & $83.9$ & $79.3$ & $67.0$ & $61.4$ \\ 
				P-1 & Meta-BIN~\cite{choi2021meta} & CVPR2021 & $57.9$ & $48.4$ & $68.6$ & $59.9$ & $87.0$ & $81.3$ & $71.2$ & $63.2$\\ 
				& DTIN-Net~\cite{jiao2022dynamically} & ECCV2022 & $60.6$ & $51.8$ & $70.7$ & $62.9$ & $\mathit{87.2}$ & $\mathit{81.8}$ & $72.8$ & $65.5$\\
				& MDA~\cite{ni2022meta} & CVPR2022 & $\mathit{62.9}$ & $\mathbf{61.2}$ & $\mathit{71.7}$ & $\mathit{63.5}$ & $84.4$ & $80.4$ & $\mathit{73.0}$ & $\mathbf{68.4}$\\
				\cline{2-11}
				& 3D-VAN & - & $\mathbf{64.6}$ & $\mathit{56.5}$ & $\mathbf{72.7}$ & $\mathbf{64.9}$ & $\mathbf{87.7}$ & $\mathbf{82.8}$ & $\mathbf{75.0}$ & $\mathit{68.1}$\\ 
				\hline \hline
				& SNR~\cite{jin2020style} & CVPR2020 & $41.3$ & $30.4$ & $65.0$ & $55.1$ & $\mathit{91.9}$ & $\mathit{87.0}$ & $66.1$ & $57.5$ \\ 
				& DMG-Net~\cite{bai2021person30k} & CVPR2021 & $47.2$ & $37.3$ & $70.9$ & $62.3$ & $88.2$ & $83.0$ & $68.8$ & $60.9$ \\ 
			    P-2 & RaMoE~\cite{dai2021generalizable} & CVPR2021 & $53.9$ & $43.4$ & $\mathit{72.2}$ & $63.4$ & $\mathbf{92.3}$ & $\mathbf{88.4}$ & $72.8$ & $65.1$ \\ 
				& DTIN-Net~\cite{jiao2022dynamically} & ECCV2022 & $\mathit{58.4}$ & $\mathit{49.4}$ & $71.9$ & $\mathit{64.0}$ & $89.2$ & $85.3$ & $\mathit{73.2}$ & $\mathit{66.2}$\\
				\cline{2-11}
				& 3D-VAN & - & $\mathbf{61.2}$ & $\mathbf{53.0}$ & $\mathbf{76.2}$ & $\mathbf{67.9}$ & $88.1$ & $83.0$ & $\mathbf{75.2}$ & $\mathbf{68.0}$\\ 
				\hline
		\end{tabular}
		}
	\end{center}
	\vspace{-8mm}
\end{table*}

\noindent where the $L^{Tri./ID}$ represent triplet loss or ID classification loss; the subscript $o/c/cls$ indicate the current loss is applied over original features ($o$), canonical view features ($c$), or class token feature ($cls$).
Particularly, to reduce computational consumption, in each training batch, we randomly sample a single canonical view to map the original images into canonical view images.
During the test stage, for each original image, we use all canonical views to map it into four canonical view images.
After that, we independently extract the features of all canonical view images and then concatenate them together with the original image feature for evaluation.



\section{Experiments}

\subsection{Implementation Details and Evaluation Setting}~\label{sec:IS}

\vspace{-4mm}
\noindent\textbf{Dataset.} In this paper, we utilize Market1501 (M)~\cite{zheng2015scalable}, Cuhk02 (C2)~\cite{li2013locally},  Cuhk03 (C3)~\cite{li2014deepreid}, CuhkSYSU (CS)~\cite{xiao2016end}, MSMT17 (MS)~\cite{wei2018person}, DukeMTMC-ReID (D)~\cite{zheng2017unlabeled}, VIPeR (V)~\cite{gray2008viewpoint}, GRID (G)~\cite{loy2009multi} and QMUL i-LIDS (Q)~\cite{zheng2009associating}  to evaluate our model.
Besides, in this work, we employ a newly proposed surveillance-to-aerial ReID benchmark named PRSA-1060~\cite{wang2022PRSA} (SA).
In the SA, all pedestrian instances are simultaneously captured by two surveillance cameras and one UAV camera, which is effective in validating the robustness of ReID models facing great viewpoint divergence.

\noindent{\bf Settings.} 
In this work, we follow previous work~\cite{jiao2022dynamically} to adopt two multiple-source domain generalization protocols and one single-source domain generalization protocol.
In Protocol-1, all the images in M, D, C2, C3, and CS are used for training, while the V, G, and Q serve as evaluation sets. 
In Protocol-2, all the images in M, D, C3, and MS are used for training, while the test set is the same as in Protocol-1. 
In Protocol-3, we train ReID models on the training set of M (D) and test it on the test set of D (M) or SA.

Besides, we also propose a novel camera-view generalization ReID setting (CVG-ReID).
Under the CVG-ReID, we modify M (D) by discarding half of the camera views in the training set while keeping them in the test set to simulate unseen camera views.
Here, we need to clarify that the original training set and the test set of M (D) are collected under totally the same cameras. 
Since the other domain factors, \eg, season, between preserved cameras and the unseen camera views are almost the same, in the CVG-ReID, we could use the modified M (D) to especially evaluate the robustness of ReID models toward novel camera views.
The detail of CVG-ReID is given in section C in the Appendix.



\noindent{\bf Implementation Details.}
In our experiments, the texture mapping module R-Texformer is trained on the source domains under Protocol-1 and Protocol-2.
Under Protocol-3 and our CVG-ReID, to avoid inadequate training data in a single-source domain causing over-fitting to R-Texformer, we train the R-Texformer on a large unlabelled human dataset LUPerson~\cite{fu2021unsupervised} and two human snapshot datasets~\cite{alldieck2018video, peng2021neural}.
In our experiments, the posture of all 3D meshes is unified to a walking posture.
For our 3D-VAN, we select ResNet-50~\cite{he2016deep} pre-trained on ImageNet~\cite{deng2009imagenet} as the backbone model.
Under all Protocols, we train our model for 120 epochs.
The learning rate is initialized to $1.0\times10^{-4}$ and divided by $10$ at the $40$th and $90$th epochs, respectively.
Random flipping and color jittering are employed for data augmentation. 
During the training and test stage, the image sizes of original images and canonical view images are both set to $256 \times 128$.
The CMC and mAP metrics are employed for evaluation. All results given in this paper are the mean of two repeated experiments.

\subsection{Comparison between Our 3D-VAN with Other State-of-the-art Algorithms}~\label{sec:sota} 

\vspace{-4mm}
\noindent{\bf Multiple-Source Domain Generalization.} We first compare our 3D-VAN with other state-of-the-art DG-ReID methods under multiple-source domain settings. The comparison results are given in Table~\ref{tab:multiple}.
Among these compared methods, the DTIN-Net~\cite{jiao2022dynamically} is a typical normalization-based method, which attempts to enhance the generalization capability of ReID models by eliminating style interference.
Compared with the DTIN-Net, our 3D-VAN achieves significantly better performance, \ie, average $2.2\%$ mAP under Protocol-1 (P-1).
The reason for the performance advantage of our 3D-VAN may be that our model can more effectively address the degradation caused by unseen camera views in cross-domain settings.
Compared with other methods, only RaMoE~\cite{dai2021generalizable} and MDA~\cite{ni2022meta} perform slightly better than our 3D-VAN on their especially suitable dataset (i-LIDs or GRID).
Nevertheless, our 3D-VAN outperforms the best competitors by $2.0\%$ mAP on average under both P-1 and P-2, which demonstrates the superiority of our model.

\noindent{\bf Single-Source Domain Generalization.} To further evaluate our 3D-VAN, we also compare it with state-of-the-art methods under single-source domain generalization settings.
In this part, in addition to generalizing our model between surveillance-based benchmarks (Market and Duke) as in~\cite{jiao2022dynamically}, we also propose to generalize it from surveillance-based datasets to aerial imagery~\cite{wang2022PRSA}, which could reflect the robustness of ReID models toward viewpoint variation intuitively.
From the comparison results given in Table~\ref{tab:single} we can find that our 3D-VAN outperforms all compared methods, \ie, averagely $3.1\%$ mAP better than the best competitors.
Particularly, under the surveillance-to-aerial setting, our 3D-VAN outperforms the normalization-based method DTIN-Net~\cite{jiao2022dynamically} by a large margin, termed $3.7\%$ mAP on average.
The reason for the performance gap could be that our 3D-VAN is more effective in addressing the viewpoint variation from surveillance to aerial imagery.

\begin{table*}
	\begin{center}
\caption{The comparison between our 3D-VAN and other methods under single-source domain generalization protocol. The ``S-t-S'' and ``S-t-A'' respectively indicate surveillance-to-surveillance generalization and surveillance-to-aerial generalization settings. 
}
	\label{tab:single}
	\vspace{-1mm}
	\resizebox{0.75\textwidth}{!}{ 
			\begin{tabular}{c| c| c | c  c c| c  c c }
				\hline
		         \multirow{2}*{Methods} & \multirow{2}*{Reference} & \multirow{2}*{Setting}  & Training & \multicolumn{2}{c|}{Test: Duke} & Training & \multicolumn{2}{c}{Test: Market} \\
				  & & & Set & mAP & CMC-1 & Set & mAP & CMC-1 \\
				\hline \hline
				QAConv~\cite{liao2020interpretable} & ECCV2020 & S-t-S & Market & $28.7$ &  $48.8$ & Duke & $27.2$ & $58.6$ \\
				SNR~\cite{jin2020style} & CVPR2020 & S-t-S & Market & $33.6$ & $55.1$ & Duke & $33.9$ & $66.7$\\
				MetaBIN~\cite{choi2021meta} & CVPR2021 & S-t-S & Market & $33.1$ & $55.2$ & Duke & $35.9$ & $69.2$\\
				MDA~\cite{ni2022meta} & CVPR2022 & S-t-S & Market & $34.4$ & $56.7$ & Duke & $38.0$ & $70.3$\\
				\hline
				3D-VAN & - & S-t-S & Market & $\mathbf{36.6}$ & $\mathbf{58.1}$ & Duke & $\mathbf{41.6}$ & $\mathbf{72.5}$\\
				\hline
		         \multirow{2}*{Methods} & \multirow{2}*{Reference} & \multirow{2}*{Setting} & Training & \multicolumn{2}{c|}{Test: PRSA} & Training & \multicolumn{2}{c}{Test: PRSA} \\
				 & & & Set & mAP & CMC-1 & Set & mAP & CMC-1 \\
				\hline \hline
				SNR~\cite{jin2020style} & CVPR2020 & S-t-A & Market & $34.6$ & $58.6$ & Duke & $34.3$ & $58.9$\\
				MetaBIN~\cite{choi2021meta} & CVPR2021 & S-t-A & Market & $37.6$ & $60.8$ & Duke & $36.4$ & $60.5$\\
				DTIN-Net~\cite{jiao2022dynamically} & ECCV2022 & S-t-A & Market & $38.7$ & $60.3$ & Duke & $35.6$ & $59.0$\\
				\hline
				3D-VAN & - &S-t-A & Market & $\mathbf{41.5}$ & $\mathbf{66.3}$ & Duke & $\mathbf{40.2}$ & $\mathbf{62.5}$\\
				 \hline
		\end{tabular}
		}
	\end{center}
	\vspace{-7mm}
\end{table*}

\begin{table}[t]
	\caption{ The effectiveness of our designed components, namely, the 3D reconstruction and alignment component (``3D'') and the transformer-based fusion component (``Trans''). The ``Asym.'' is the asymmetric access restriction in the fusion module that avoids the information transferred from the canonical view image to the original image. The experiments are based on Protocol-1. In particular, without using the ``3D'' component to generate canonical view images, the only employed ``Trans'' will be used as a refinement module to improve the original image features. 
	}
	\vspace{-4mm}
        \setlength\tabcolsep{4pt}
	\begin{center}
		\resizebox{0.48\textwidth}{!}{
			\begin{tabular}{c | c c c| c c | c c }
				\hline
				& \multirow{2}*{3D}  & \multirow{2}*{Trans} & \multirow{2}*{Asym.}  & \multicolumn{2}{c|}{GRID} & \multicolumn{2}{c}{VIPeR} \\
				& ~ & ~ & ~& mAP & CMC-1 & mAP & CMC-1 \\
				\hline \hline
				Baseline & $\times$  & $\times$ & $\times$ &  $56.8$ & $46.7$ & $67.8$ & $58.9$ \\
				+ 3D & $\checkmark$  & $\times$ & $\times$ & $58.6$ & $48.1$ & $69.2$ & $60.0$ \\
				+ Trans. & $\times$  & $\checkmark$ & $\times$ &  $58.3$ & $48.0$ & $68.9$ & $59.2$ \\
				\hline
				3D-VAN* & $\checkmark$ & $\checkmark$  & $\times$  &  $62.8$ & $54.3$ & $71.3$ & $63.6$  \\ 
				3D-VAN & $\checkmark$ & $\checkmark$  & $\checkmark$  &  $64.6$ & $56.5$ & $72.7$ & $64.9$  \\ 
				\hline
		\end{tabular}}
	\end{center}
	\vspace{-8mm}
	\label{tab:com}
\end{table}

\subsection{Ablation Studies}\label{sec:abl}
To clarify, in the rest of this paper, the ``Baseline'' model indicates a ResNet-50 model inserted with SNR modules and trained with triplet loss and ID classification loss. More ablation experiments, \eg, the effect of multiple canonical view feature fusion, are given in Section D in the Appendix.

\noindent{\bf The Effectiveness of Each Component in Our 3D-VAN.}
To explore the effectiveness of each component in our 3D-VAN, namely, the 3D human reconstruction and alignment component (``3D'') and transformer-based feature fusion component (``Trans''), we gradually insert them into the baseline model and compare the performance improvement under Protocol-1.
In particular, without the ``3D'' component to generate canonical view images, the only employed transformer module in ``+ Trans''  will be used as a feature refinement module to improve the original image features.
Besides, in this part, we also validate the effectiveness of asymmetric access restriction (Asym.) we adopted in the feature fusion module, which avoids the information transferring from canonical view features to original features.

The experimental results are given in Table~\ref{tab:com}, in which the ``+ 3D''  could be regarded as an ensemble model that concatenates features of canonical view image and original image together for evaluation yet only brings a slight $1.6\%$ mAP improvement on average to the baseline model.
The reason for the limited improvement could be the unsatisfactory quality of generated canonical view images.
Besides, the ``+ Trans'' is the version that inserts the transformer modules~\cite{wu2021cvt} to the baseline model, which only brings an average $1.3\%$ mAP improvement to the baseline model.
\begin{figure}[t]
	\begin{center}
		\includegraphics[width=1.0\linewidth]{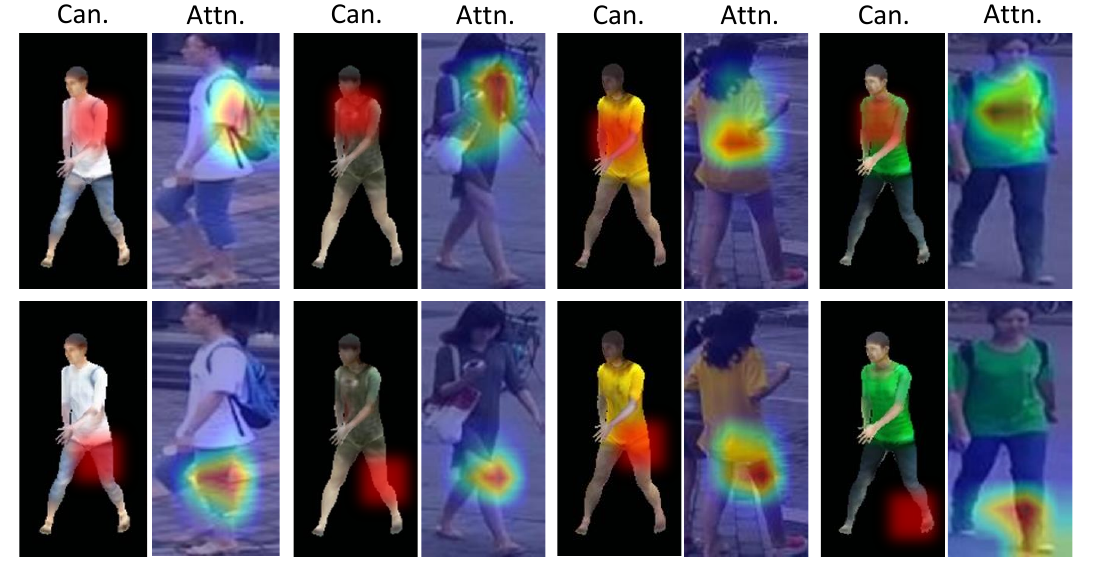}
	\end{center}
    \vspace{-6mm}
	\caption{
    The visualization of attention maps (Attn.) generated by our feature fusion module. The marked regions in canonical view images (Can.) are used as queries to apply attention to original images. We can find that our feature fusion module could effectively align the textural clues between original images and canonical view images.
		}
	\label{fig:visualization}
	\vspace{-6mm}
\end{figure}
It means the superiority of our 3D-VAN is not simply caused by using the effective transformer structure.
In fact, the effects of our ``3D'' component and feature fusion module are closely knitted.
As shown in Figure~\ref{fig:visualization}, our fusion module could effectively align and fuse the texture clues from original images to canonical view images, which implicitly compensates for the damaged details suffered by the ``3D'' component, resulting in a significant improvement for our final model ``3D-VAN'' that uses both components simultaneously.
In addition, comparing the results of ``3D-VAN*'' and ``3D-VAN'', we can find that the access restriction in the fusion module brings an average $1.6\%$ mAP improvement. The reason for the performance gain could be that this design avoids the damaged textural information transferred to the original image features.



\begin{figure}[t]
	\begin{center}
		\includegraphics[width=1.0\linewidth]{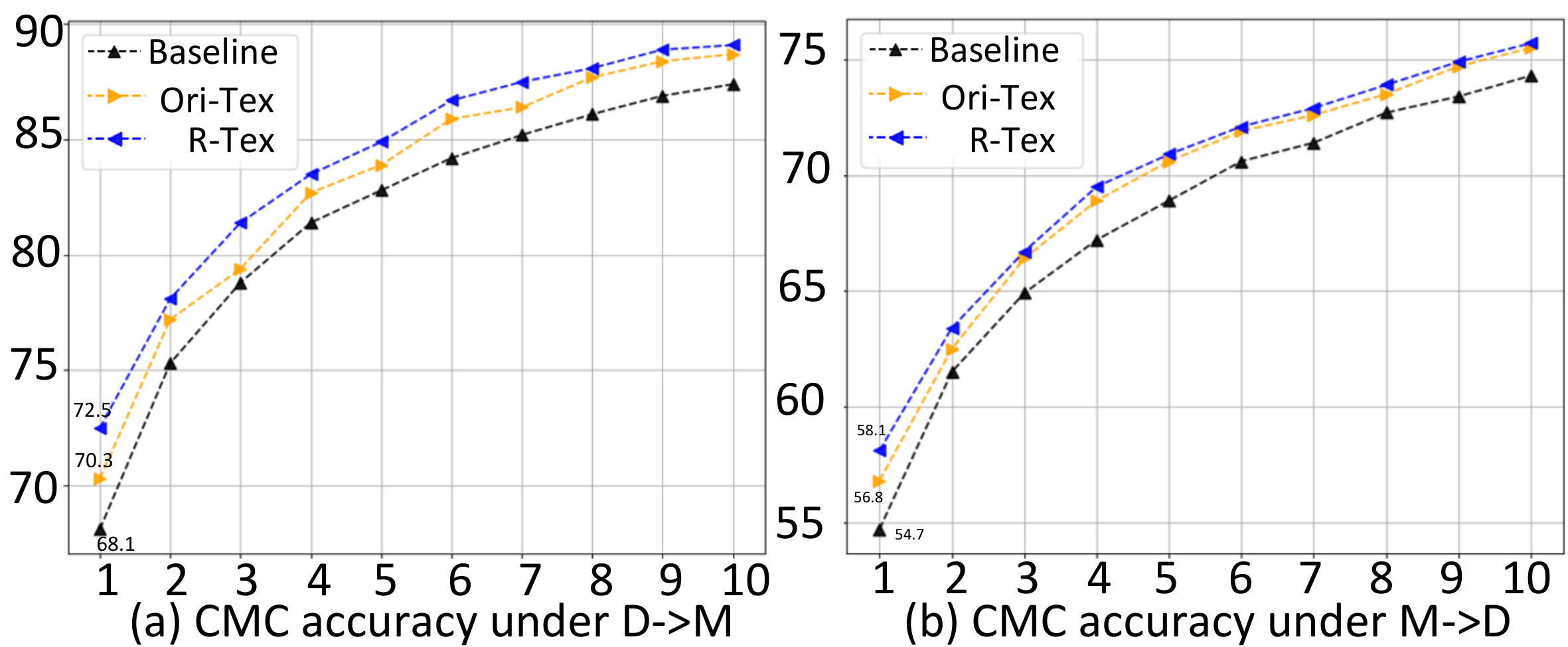}
	\end{center}
    \vspace{-6mm}
	\caption{The effectiveness of the modifications we apply over Texformer. We compare the CMC-1 to CMC-10 accuracy of the 3D-VAN model employing different versions of Texformer under Protocol-3. The ``Ori-Tex'' indicates the 3D-VAN using the original Texformer. The ``R-Tex'' is the 3D-VAN using R-Texformer.
		}
   \vspace{-2mm}
	\label{fig:Mends}
\end{figure}

\noindent{\bf The Effect of Modifications over Texformer.}
In this part, we validate the effort of our modifications over Texformer. 
To do so, we respectively use an original Texformer model and a modified R-Texformer to support the training of our 3D-VAN.
The experiments are based on Protocol-3, in which we train the 3D-VAN on Market (Duke) and test it on Duke (Market). 
The versions of 3D-VAN using the original Texformer and R-Texformer are denoted as ``Ori-Tex'' and ``R-Tex'', the performance of which is given in Figure~\ref{fig:Mends}.
From the results, we can find that our refinements lead to encouraging improvements, specifically, an average $1.8\%$ CMC-1 improvement from ``Ori-Tex'' to ``R-Tex''.
The performance gain indicates that our refinements over Texformer indeed help it produce high-quality textural predictions to enhance the identifiability of canonical view images.
In Figure C in the Appendix, we give some visualization cases to illustrate how our refinements improve the texture quality.

\begin{table}[t]
	\caption{
    Comparison between our 3D-VAN and other methods under CVG-ReID. The ``S.''  denotes that this method is designed for the standardized ReID setting. The ``D.G.'' denotes that this method is designed for the domain generalization ReID setting.
	}
	\vspace{-6mm}
	\begin{center}
		\resizebox{0.48\textwidth}{!}{
			\begin{tabular}{c | c | c c | c c }
				\hline
				   & \multirow{2}*{Setting} & \multicolumn{2}{c|}{Market} & \multicolumn{2}{c}{Duke} \\
				& ~ & mAP & CMC-1 & mAP & CMC-1 \\
				\hline \hline
				PCB~\cite{sun2018beyond} & S. & $52.7$ & $74.3$ & $47.0$ & $64.4$ \\
				Pyramid~\cite{zheng2019pyramidal} & S. & $56.9$ & $73.8$ & $49.5$ & $64.5$ \\
				ABDNet~\cite{chen2019and-net} & S. & $60.8$ & $80.3$ & $51.2$ & $67.4$ \\
				CAL~\cite{rao2021counterfactual} & S. & $67.2$ & $83.1$ & $56.4$ & $72.9$ \\
				\hline
				SNR~\cite{jin2020style} & D.G. & $64.9$ & $85.0$ & $54.4$ & $73.0$ \\
				DTIN-Net~\cite{jiao2022dynamically} & D.G. & $66.9$ & $84.8$ & $56.7$ & $74.1$ \\
				\hline 
				3D-VAN & D.G.  & $\mathbf{71.2}$ & $\mathbf{87.1}$ & $\mathbf{59.8}$ & $\mathbf{76.0}$ \\
				\hline
		\end{tabular}}
	\end{center}
	\label{tab:cvg}
	 \vspace{-8mm}
\end{table}

\noindent{\bf Camera-view Generalization Capability of 3D-VAN.}
To intuitively validate the robustness of our 3D-VAN toward viewpoint variation, in this part, we compare 3D-VAN with other ReID methods under our proposed CVG-ReID setting.
The results are given in Table~\ref{tab:cvg}, in which the ``S.'' indicates models designed for standardized ReID setting~\cite{sun2018beyond,rao2021counterfactual} (training and test data are collected from totally the same domains and cameras), while the ``D.G.''  indicates models designed for domain generalization ReID setting.
From these results, we could get three important findings.
Firstly, the advanced ``S.'' method CAL~\cite{rao2021counterfactual} still has an average $3.7\%$ mAP gap with our 3D-VAN when facing unseen test cameras, which could indicate the unpredictable camera views in novel domains is indeed a crucial factor limiting the generalization of ReID models.
Secondly, under the CVG-ReID, normalization-based DG-ReID methods (SNR and DTIN-Net) only achieve comparable performance to the standardized ReID methods.
The reason could be that the variation of domain factors other than camera views, like the season, in our CVG-ReID is subtle, and existing normalization-based methods are ineffective in tackling viewpoint changes.
Finally, the superiority of our 3D-VAN over all other compared methods indicates that our model could effectively solve the degradation caused by unpredictable viewpoint variation in novel domains.



\begin{figure}[t]
	\begin{center}
		\includegraphics[width=1.0\linewidth]{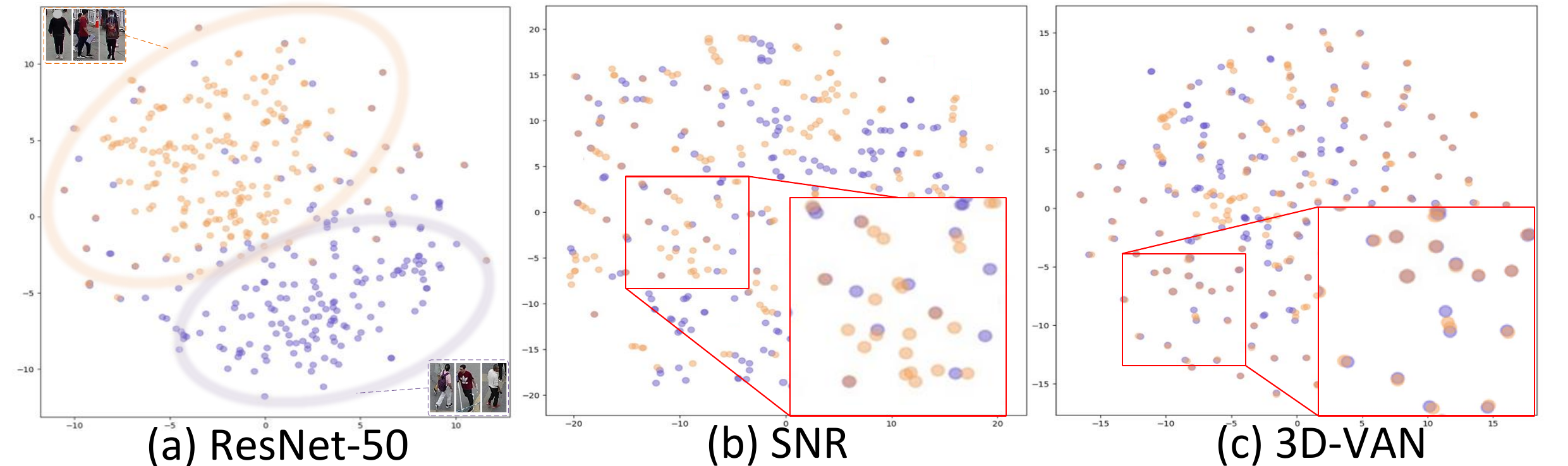}
	\end{center}
    \vspace{-4mm}
	\caption{t-SNE visualizations. 
	    Please zoom in for a better view. We select 200 pedestrians from the PRSA dataset. For each pedestrian, we sample two images, one from a surveillance camera (orange points) and another one from a UAV camera (violet points).
		}
   \vspace{-4mm}
	\label{fig:tsne}
\end{figure}

\noindent{\bf T-SNE Visualization Results.}
In Figure~\ref{fig:tsne}, we give the t-SNE results of a ResNet-50~\cite{he2016deep} model, a SNR~\cite{jin2020style} model, and our 3D-VAN. Here, we train all these models with triple loss and ID classification loss on DukeMTMC~\cite{zheng2017unlabeled} dataset and use them to extract features on PRSA~\cite{wang2022PRSA} dataset. Specifically, we select 200 pedestrians from the PRSA dataset and sample two images for each selected pedestrian, one from a surveillance camera (orange points) and the other one from a UAV camera (violet points).
From Figure~\ref{fig:tsne} (a), we can find that the viewpoint divergence across cameras brings a significant feature distribution gap to ResNet-50, which may affect the identification. For SNR~\cite{jin2020style}, even though it breaks the distribution gap between different camera views via instance normalization, it fails to match features of the same pedestrian into compact sub-clusters.
From Figure~\ref{fig:tsne} (c), we can find that through the explicit viewpoint alignment, our 3D-VAN could achieve pedestrian matching across novel cameras and produce relatively compact sub-clusters.

\section{Conclusion}
In this work, we identify that viewpoint variation is one of the major factors affecting the generalization of ReID methods.
Hence, we propose a 3D-VAN model which uses the 3D space as a bridge to calibrate pedestrian images from arbitrary domains into canonical view images.
Besides, we also design a transformer-based feature fusion module to align and fuse the texture clues from the original image to the canonical view images, which compensates for the information loss caused by imperfect 3D reconstruction technology and achieves better feature presentations for ReID.

\clearpage
%
%
\bibliographystyle{splncs04}
\bibliography{main}
\end{document}